\documentclass[a4paper, 10pt, conference]{ieeeconf}      

\IEEEoverridecommandlockouts                              

\usepackage{graphics} 
\usepackage{epsfig} 
\usepackage{amsmath} 
\usepackage{amssymb}  
\usepackage{url}
\usepackage{hyperref}
\usepackage{multirow}
\hypersetup{
    colorlinks=true,
    linkcolor=blue,
    filecolor=magenta,      
    urlcolor=blue,
}

\title{\LARGE \bf
DDD20 End-to-End Event Camera Driving Dataset: Fusing Frames and Events with Deep Learning for Improved Steering Prediction
}

\author{Yuhuang Hu$^{1}$, Jonathan Binas$^{1,2}$, Daniel Neil$^{1,3}$, Shih-Chii Liu$^{1}$ and Tobi Delbruck$^{1}$
\thanks{$^{1}$Yuhuang Hu, Shih-Chii Liu and Tobi Delbruck are with the Institute of Neuroinformatics, University of Z\"urich and ETH Z\"urich, Switzerland
        {\tt\small \{yuhuang.hu, shih, tobi\}@ini.uzh.ch}}%
\thanks{$^{2}$Jonathan Binas is now with MILA, University of Montreal, Canada
        }%
\thanks{$^{2}$Daniel Neil is now with Benevolent AI, NY, USA
        }%
        \thanks{This work was funded by Samsung via the Neuromorphic Processor Project (NPP),  the Swiss National Competence Center in Robotics (NCCR Robotics), and the EU projects \href{http://www.seebetter.eu/}{SEEBETTER} and \href{http://www.visualise-project.eu/}{VISUALISE}.}
}

\begin{document}

\maketitle
\thispagestyle{empty}
\pagestyle{empty}

\begin{abstract}

Neuromorphic event cameras are useful for 
dynamic vision problems under difficult lighting conditions.
To enable studies of using event cameras in automobile driving applications, 
this paper reports a new end-to-end driving dataset called DDD20.
The dataset was captured with a DAVIS camera that concurrently streams both dynamic vision sensor 
(DVS) brightness change events and active pixel sensor (APS) intensity frames.
DDD20 is the longest event camera end-to-end driving dataset to date with 51h of DAVIS event+frame camera and vehicle human control data collected from
 4000\,km of highway and urban driving under a variety of lighting conditions.
Using DDD20, we report the first study of 
fusing brightness change events and intensity frame data using
a deep learning approach to predict 
the instantaneous human steering wheel angle.
Over all day and night conditions, the explained variance for human steering prediction
from a Resnet-32 is significantly better from the fused \mbox{DVS+APS} frames (0.88) than
using either DVS (0.67) or APS (0.77) data alone.

%

\end{abstract}

\section{INTRODUCTION}

Recent advances in autonomous driving~\cite{Bojarski:2016,Chen:2015,Sharifzadeh:2016,Xu:2016, adas:review:Grigorescu:2019} have been fueled by modern deep learning methods, whereby driving controllers are typically trained on extensive datasets of
real-world recordings and simulated environments~\cite{Cordts:2016,Geiger:2012,Maddern:2017,santana_learning_2016,Yu:2017}.
The availability of these datasets together with advances in deep
learning has enabled improvements in computer vision 
technologies that are essential to the success of autonomous
driving, such as semantic segmentation~\cite{Zhao:2017}, 
object detection, tracking~\cite{Ren:2015}, and motion
estimation~\cite{Kendall:2015}.

\begin{figure}[t]
	\centering
    \includegraphics[width=0.57\columnwidth]{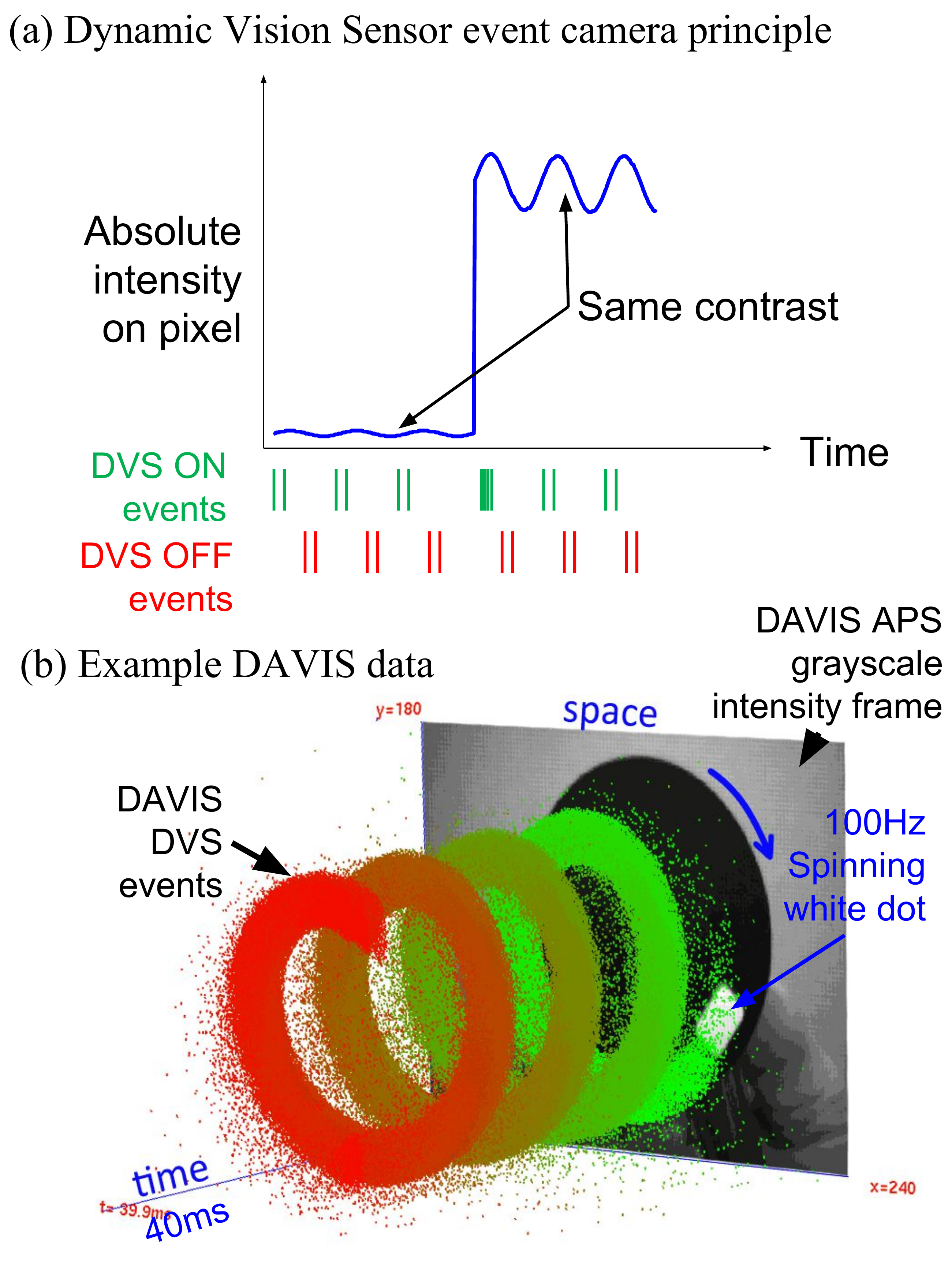}
    \caption{(a) Concept of DVS event camera pixel response; (b) DAVIS frame + event camera data from 100\,Hz spinning dot.}\label{fig:davis}
\end{figure}

Self-driving vehicles must operate under a wide range of lighting conditions, 
and thus it is crucial that employed vision sensors offer high dynamic range 
and high sensitivity, enabling short exposure 
times to minimize motion blur. Event cameras such as Dynamic Vision Sensors (DVS)~\cite{Lichtsteiner:2008} can offer advantages under conditions that are difficult for conventional cameras. 
In contrast to regular-sampled, frame-based cameras, \emph{event cameras} 
produce a stream of asynchronous \textit{timestamped address events} that are 
triggered by local brightness (log intensity) changes at individual pixels. 
Fig.~\ref{fig:davis}(a) shows the principle of the DVS pixel response. 
The DVS responds the same way to equal contrast variations 
(typically caused by scene reflectance changes) independent of absolute intensity.
The local instantaneous gain control enables a broader
dynamic range than conventional cameras (120dB vs.\,60dB)  for handling uncontrolled lighting conditions. 
These events are transmitted off-chip with submillisecond latency. 
Each event includes
the pixel coordinates, the sign of the brightness change, and the microsecond timestamp. 
The asynchronous nature of DVS events reduces the latency and 
bandwidth requirements of the system, enabling robots with millisecond response times at low average CPU load. 
Automotive cameras require special pixel architectures 
 to minimize frame-to-frame aliasing of pulse-width-modulated LED light sources 
 like car taillights and traffic sign light sources. 
The high temporal resolution of the DVS events  
enables accurate CNN-based~\cite{EVFLOWNET:Zhu:2018} 
and sub-ms hardware-based~\cite{ABMOF-Liu2018-yn} optical flow estimation 
and flashing
light source detection and tracking~\cite{Conradt-Muller2011-af}.

Fig.~\ref{fig:davis}(b) shows  output from 
a next generation event camera called Dynamic and 
Active-Pixel Vision Sensor (\textbf{DAVIS})~\cite{Brandli:2014}. 
A DAVIS concurrently outputs both DVS events (the spiral cloud of 
points in Fig.~\ref{fig:davis}(b)), 
and standard global-shutter active pixel sensor (\textbf{APS}) 
intensity frames (background image in Fig.~\ref{fig:davis}(b)).
The DVS and APS pixel
circuits that share the same photodiode.
The combination of sampled analog gray values from the APS 
stream and the asynchronous, high dynamic range brightness 
change events from the DVS could make the DAVIS well-suited to 
driving applications: When the APS stream is over- or under-exposed, 
or the features are blurred or aliased, the DVS events can 
provide the missing information.

\subsection{Related work}

\cite{Moeys:2016} showed that fused
DAVIS frame+event data could drive a CNN to steer a predator robot to follow a prey robot.
It inspired us to 
investigate benefits 
that the DAVIS camera could provide for autonomous driving.
To avoid expensive data labeling as in \cite{Moeys:2016},
we followed the pioneering end-to-end (\textbf{E2E})
studies dating back to ALVINN~\cite{Pomerleau:1989, Pomerleau:1992},
and more recently comma.ai and NVIDIA~\cite{Bojarski:2016,santana_learning_2016}, where
the network directly predicts
the human's instantaneous steering wheel angle based on the appearance of the road. 

Our first dataset 
called DDD17, containing 12h of E2E labeled driving data~\cite{Binas:2017},  was used by
\cite{Maqueda:2018} 
to compare the human steering with predictions using APS frames,  
APS frame differences, and DVS frames. 
They showed that DVS frames gave better steering prediction than
 APS frames (in contrast to our findings in Sec.~\ref{subsec:steeringpredictionresults}),
 and better prediction than APS frame differences, however, they did not evaluate
 the benefit of fusing DVS and APS data. 
\cite{Maqueda:2018} also found that ResNet 
CNN architectures are well 
suited to the steering prediction problem, and that
a DVS `frame' duration of 50ms produced the best predictions.

DDD17 was limited in road types, weather and daylight conditions. 
Since then, MVSEC~\cite{MVSEC:Zhu:2018b}, DET~\cite{DET:Cheng:2019}, 
Event Camera Driving Sequences~\cite{Rebecq19pami},
and GET1~\cite{GEN1:automotive:Tournemire:2020} datasets have been 
released. These datasets contain useful driving data with various label types, 
but none of them are E2E labeled with human driving.

\subsection{DDD20}

To allow more extensive E2E studies, we expanded the 12\,h DDD17 with 
an additional 39\,h of data, giving the new DDD20 dataset.
It has a total of 1.3\,TB of data with 51\,h of recordings collected from a $346\times 260$-pixel 
DAVIS346 camera, along with car parameters such as steering wheel angle.
DDD20 has recordings of 
rural highway driving under difficult sunlight glare conditions,
day and night driving in urban Los Angeles and San Diego, and
multiple repeats of the same sections of
mountain highway driving (along the Colorado Lizard Head Pass highway and California Angeles Crest Highway) 
during daylight, evening, and night. 
Section~\ref{sec:method} contains details of the dataset.

\begin{figure*}[htb]
    \vskip3ex
    \centering
    \includegraphics[width=0.85\textwidth]{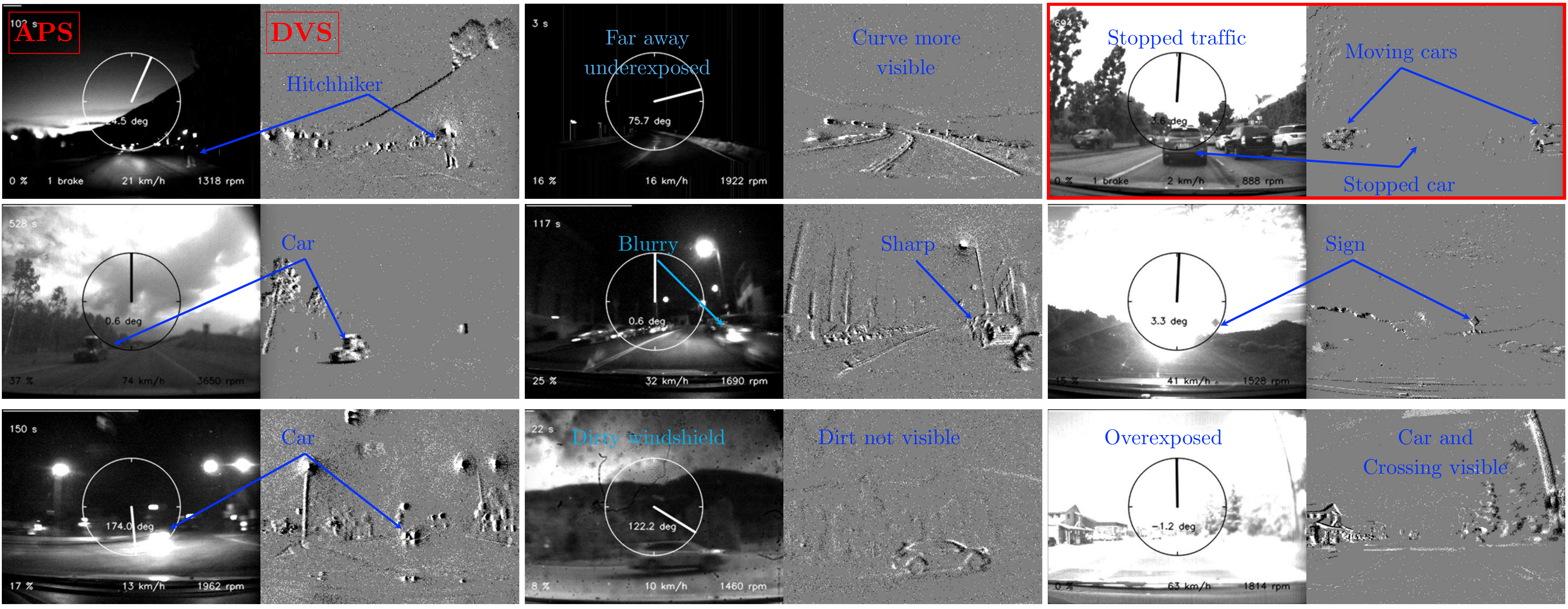}
    \caption{DDD20 sample data. Each pair of frames show APS (left) and DVS (right) data. DVS signed event histogram frame durations are 25\,ms. APS frame exposure duration varies with illumination from 50\,us to about 200\,ms.}\label{fig:sampleimages}
\end{figure*}

Fig.~\ref{fig:sampleimages} shows 
examples of how the APS and DVS streams 
from DDD20 complement each other.
For the frame pair outlined in red, 
the stopped car is invisible in the 
 DVS frame, but cars in other lanes
pop out in the DVS frame because of their motion.
For some scenes (e.g. left middle), the road edge is not visible in the DVS frame because the car is
driving straight along the road. In others (e.g. top middle), the upcoming curve is visible in
the DVS frame because
the car is approaching the curve.
In many scenes the APS frame 
is underexposed, overexposed, or motion blurred, 
but in the DVS frame, the object is still visible 
because of its superior dynamic range and quicker response. 
A properly trained network should take advantage of this complementary APS and DVS information.
We demonstrate this using a network for steering prediction in Sec.~\ref{sec:experiments}.

DDD20 includes E2E vehicle control and diagnostics 
data to allow studies of the effectiveness
of the DAVIS camera compared to standard grayscale image sensors.
It does not contain LIDAR, radar, and other 
sensors necessary for a complete 
ADAS solution.

The main contributions of this paper are 
\begin{enumerate}
    \item The DDD20 dataset, with the methods and 
software used for the dataset collection. (Sec.~\ref{sec:method}).
    \item The use of DDD20 for the first study of fusing of APS and DVS data 
for steering prediction (Sec.~\ref{sec:exps}). 
    In contrast to~\cite{Maqueda:2018}, we find that APS 
    frames produce better steering prediction results
    than DVS frames alone. In addition, we show that fusing APS and DVS data 
    improves the predictions by a significant margin over either modality by itself. 
\end{enumerate}

\section{METHODS} \label{sec:method}

The `DAVIS Driving Dataset 2020' (DDD20) dataset 
 will be released at
 {\small
 \url{http://sensors.ini.uzh.ch/databases.html}
 }.
This section describes the DAVIS camera and how we collected the dataset.

\begin{figure}[htb]
    \centering
    \includegraphics[width=0.7\linewidth]{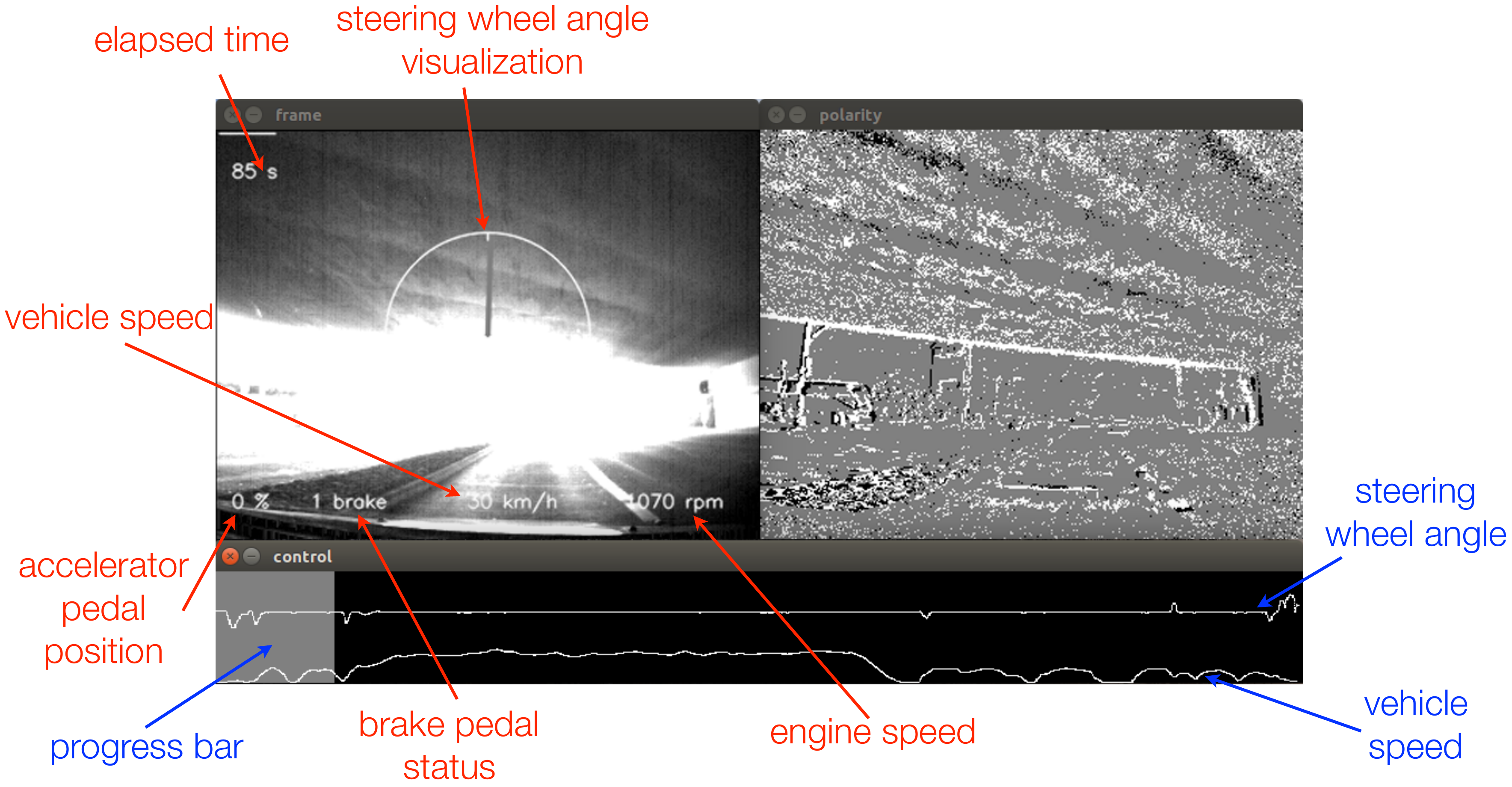}
    \caption{Example DDD20 recording visualized by the viewer application. 
    The top panels display APS (top-left) and DVS (top-right) frames. 
    The bottom panel controls and shows the status of the recording. 
    Various vehicle data fields are presented as well. 
    In this example, the APS frame suffers from over and underexposure, 
    while the DVS frame still exposes these details.}\label{fig:viewer}
\end{figure}

\subsection{DAVIS camera setup}
\label{subsec:davisdata}
Camera input was captured from our DAVIS346B~\cite{Taverni:2018}. 
It produces
both DVS events and APS frames that are 
concurrently captured from the same optics. 
Each pixel uses the same photodiode for producing DVS and APS outputs simultaneously.
The DAVIS346B with 346$\times$260 pixels is similar to the DAVIS240C~\cite{Brandli:2014},
but has 2.1$\times$ more pixels and includes on-chip column parallel 
analog-to-digital converters (ADCs) for an APS frame
rate of up to 50\,Hz.
The DAVIS346B 
also has buried photodiodes with microlenses 
and anti-reflection coating that increases the quantum efficiency by a factor of about 4, and 
reduces the photodiode dark current.
In the original DDD17 recordings, we used a 6\,mm lens providing a 56${^\circ}$ horizontal 
field of view (FOV). For the DDD20 recordings, we 
increased the FOV to 71${^\circ}$ to cover more road features during
turns by using a Kowa 4.5\,mm. 

The APS frame rate depends 
on the auto-exposure duration. 
In later recordings, this duration was set by an 
auto exposure algorithm that optimized the exposure time for 
the road surface in the lower third of the image. 
Thus the frame rate (and exposure duration) varies
between 8\,fps and 50\,fps.
In some recordings, the frame rate was also limited to reduce file size.
The frames were captured using the DAVIS global shutter mode to minimize motion artifacts.

The camera was mounted using a glass suction tripod fixed behind the windshield, 
just below the rear mirror, and aligned to point to the center of the hood. 
The original DDD17 Ford Mondeo dataset (see Sec.~\ref{sec:Mondeo}) used a single mounting point. 
For the Ford Focus recordings,
the sensor was mounted each day and adjusted to bring the edge of the hood 
to the bottom of the frame, centered on the road; 
thus recordings have slightly different viewpoints.
The USB3.0 camera was connected to a laptop computer and was read out using cAER\footnote{cAER-1.1.2 release: \url{https://gitlab.com/inivation/dv/dv-runtime/-/tags/caer-1.1.2}}, which streamed it 
 by local UDP to the recording framework described 
in Sec.~\ref{sect:software}.

\subsection{Vehicle control and diagnostic data collection}
\label{sec:Mondeo}
Data was acquired using a 2015 Ford Mondeo MK3 European Model for Swiss/German recordings 
and a 2016 Ford Focus for USA recordings. 
A \$130 OpenXC Ford Reference vehicle interface was 
connected to the passenger compartment OBDII port, and 
read out control and diagnostic data from the car's CAN bus.
The vehicle interface was connected to the 
laptop
USB port\footnote{OpenXC vehicle interface: \url{http://openxcplatform.com/vehicle-interface/hardware.html}}.
The vehicle interface was 
read out using the OpenXC python 
library, and
passed to the custom recording software described in Sec.~\ref{sect:software}.
Data including the examples in 
Table~\ref{tab:vehicledata} was sampled at about 10\,Hz. 

\subsection{Recording and viewing software}
\label{sect:software}
The  Python software framework
\texttt{ddd20-utils}\footnote{\url{https://github.com/SensorsINI/ddd20-utils}}
records, views and exports  recordings
(Fig.~\ref{fig:viewer}). 
Since the APS frames and DVS data are microsecond time-stamped 
on the camera using its clock
unlike the data provided by the vehicle interface,
both data streams were augmented 
with the millisecond 
system time of the recording computer so that the car and camera streams can be synchronized.
When possible, the computer time was  synchronized to a standard time server before recordings.
The streams were processed by separate threads.
Although the camera and car interface each have their own precision clocks, 
it was most straightforward to use the computer time to synchronize the two streams, 
because the vehicle sample rate is only about 10\,Hz.  
The data is stored in HDF5 containers.



\begin{table}[htb]
 \centering
 \caption{Subset of vehicle control and diagnostic data}\label{tab:vehicledata}
 \begin{tabular}{rlll}
    \hline
    \multicolumn{1}{l}{\textbf{ID}} & \multicolumn{1}{l}{\textbf{Data Field}} & \multicolumn{1}{l}{\textbf{Unit}} & \multicolumn{1}{l}{\textbf{Range}} \\ 
    \hline
    1. & accelerator pedal position & percent & 0--100\% \\
    2. & brake pedal status & binary & pressed/released \\
    3. & engine speed & rpm &  \\
    4. & headlamp status & binary & ON/OFF \\
    5. & latitude & degrees & \\
    6. & longitude & degrees & \\
    7. & odometer & km or miles & \\
    8. & steering wheel angle & degrees & up to $\pm$720$^{\circ}$ \\
    9. & transmission gear position & gear no. & 1 to 6 \\
    10. & vehicle\_speed & km/h & 0-160 \\
    11. & windshield wiper status & binary & ON/OFF \\
    \hline
 \end{tabular}
\end{table}

\subsection{Recorded DDD20 data}

The 51\,h of usable data were recorded under 
various weather, driving, road, and lighting conditions 
over about 60\,days of intermittent recording.
Recordings were started and stopped manually. 
They have durations between a few minutes and an hour, 
with a median length of about 700\,s.
The 40 recordings of the original DDD17 
were supplemented by an additional 175 recordings in DDD20.
Steering angles on straight roads were dominated by small deviations of $\pm 10{^\circ}$. 
The car speed was uniformly distributed over the range of 0--130\,km/h.

\section{EXPERIMENTS} \label{sec:exps}
\label{sec:experiments}

This section reports experiments using DDD20 
to address the question of whether
\emph{fusing} APS and DVS together provides better 
steering prediction than either single modality.

\subsection{Experiment configurations} \label{subsec:cofig}

\subsubsection{Data selection}
We selected 30 recordings from DDD20 that
covered a range of road types and lighting conditions, with 15 night and 15 day recordings
(recordings used are reported on DDD20 website). 
We manually pruned the ends where the car was pulling onto or off the road.
For each recording, we chose the 
first 70\% of the data as 
a part of the training data and the last 30\% as a part of the test data.
Then, we prepared three datasets: \texttt{Night}, \texttt{Day}, and \texttt{All}.
These datasets let us study the network's predication accuracy with different choices of sensor input (DVS+APS, \mbox{DVS-only}, and \mbox{APS-only}) under day versus night lighting conditions.

\subsubsection{Preprocessing inputs for training}

\cite{Maqueda:2018} showed that a 50\,ms DVS frame duration provided the optimum for DVS
steering prediction using our original DDD17 dataset~\cite{Binas:2017}.
Hence we used DVS frames of 2D histograms of signed ON/OFF DVS event counts accumulated for 50\,ms
to approximately match the average APS frame rate. With this integration time, motion blurring was acceptable
for normal passenger car dynamics.
The DVS histogram was then clipped at three times its standard deviation.
For APS-only prediction, we used the APS frames at their native sample rate.
When the APS frame rate was lower than 20Hz, we duplicated the APS frames. 
The resulting DVS frames and the corresponding APS grayscale frames were both rescaled 
to the range $[0, 1]$ following the procedure established in~\cite{Moeys:2016}.


Speeds below 15\,km/h were eliminated because they generally
signal when the car was exiting a parking space or turning a corner at an intersection.
When the car is stopped, 
the driver sometimes plays with the wheel for a while, leaving it then stopped at a random angle. 
This simple exclusion works well for our current steering prediction because 
 each prediction is based on the instantaneous DVS+APS frame, and it is not possible to know the intention of the driver such as when they are backing out of a space or deciding to make
a turn at a corner.

The distribution of the steering angles is unbalanced because straight driving dominates the recordings.
Therefore, we randomly pruned 70\% of those frames that have steering angles between $\pm 5^{\circ}$.
We also filtered out frames where the steering angles are 
larger than three times the standard deviation of all steering 
because these generally represent outliers such as pulling off the road. 
Pruning leaves about 50\% of the frames from the original training dataset.
In the test dataset, we only filtered out the extra large steering angle and low-speed frame outliers.

To reduce computation, the original 
APS and DVS frames were subsampled from $346\times260$ to $172\times128$ pixels where we could
still clearly see the road ahead. 
We aligned the camera and car inputs  
 using the system clock timestamp.

\subsubsection{Baseline network}
\begin{figure}[htb]
	\centering
    \includegraphics[width=0.2\textwidth]{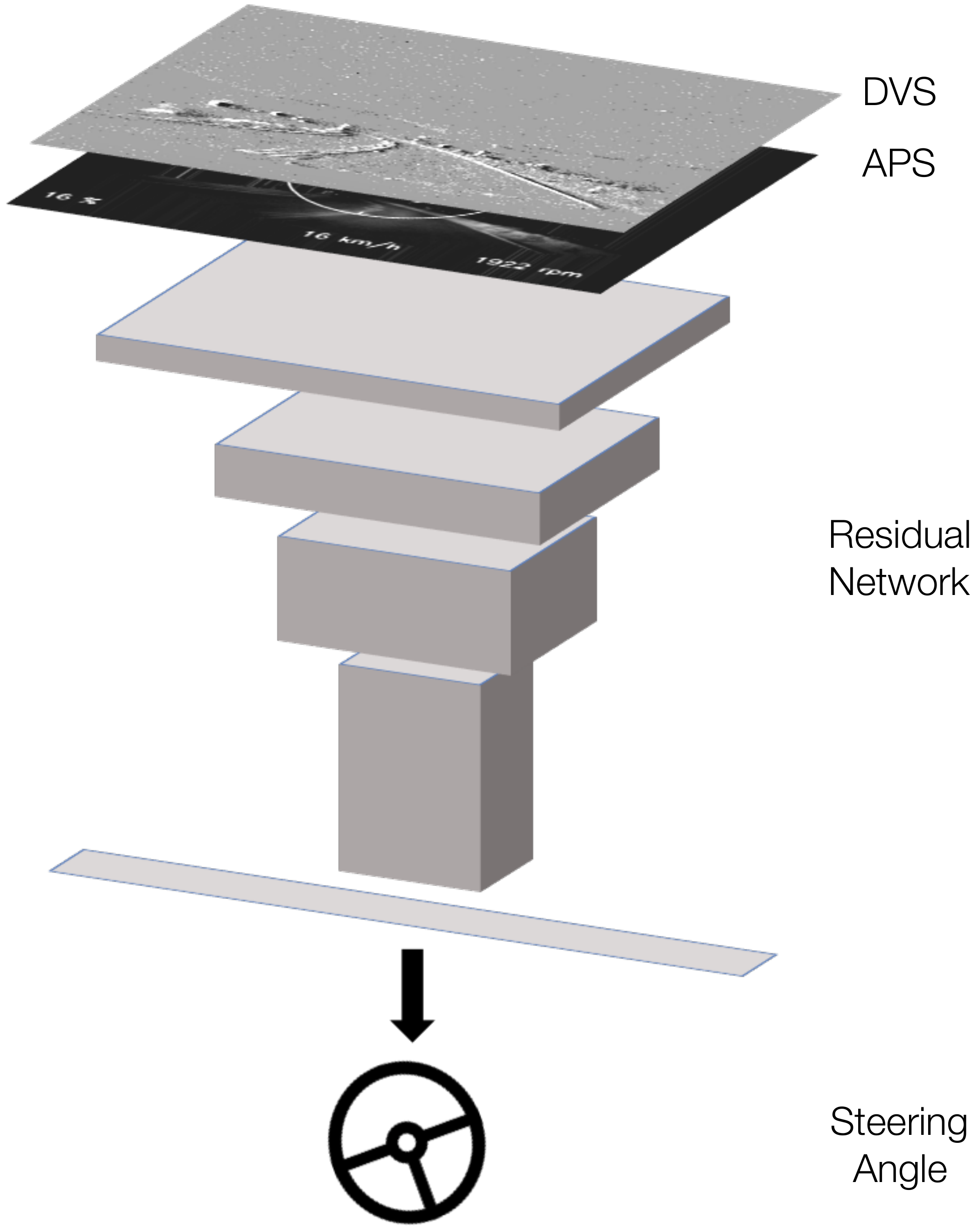}
    \caption{Illustration of the steering prediction network in this paper.}\label{fig:resnet}
\end{figure}

Based on~\cite{Maqueda:2018}, we chose the 32-layer Residual Network (\mbox{ResNet-32}) 
as the baseline network to study 
the steering angle prediction.
The configuration for the convolution layers is identical
to the one in~\cite{He:2015}. The output layer is a linear
layer that has one output for predicting the steering angle.
Fig.~\ref{fig:resnet} shows the architecture.
In the cases of \mbox{DVS-only} 
and \mbox{APS-only}, the network is trained with a one-channel input.
The network has 470k parameters and has about 400M connections.
\begin{figure*}[htb]
    \centering
    \includegraphics[width=0.7\textwidth]{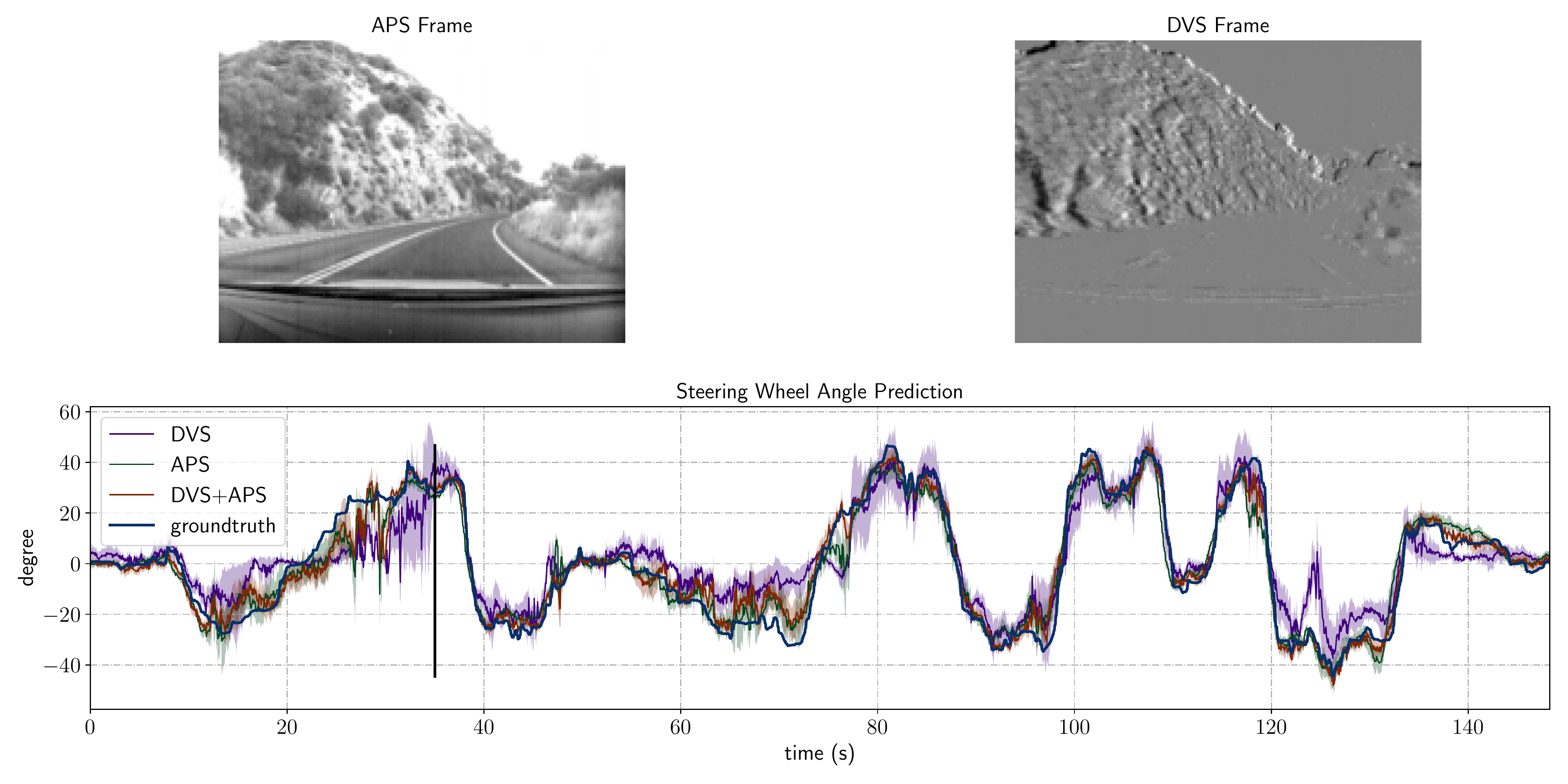}
    \caption{Experiment results for \texttt{rec1501288723.hdf5}.
    Top figures show the 700th APS and DVS frames. 
    The black vertical line in the plot also indicate the 700th frame. 
    All networks can successfully predict 
    the steering wheel angle but the DVS+APS one is most accurate.
    Shadings show $1\sigma$ standard deviation over 5 training/test runs.}\label{fig:resnet:result}
\end{figure*}

\subsubsection{Training details}
\label{sec:trainingdetails}

The weight parameters were initialized following
\cite{He:2015} by sampling from a Gaussian distribution 
$\mathcal{N}(0, 2/\sqrt{n_\textrm{in}})$, where $n_\textrm{in}$ is the number of input neurons. 
Our dataset was large enough so that we did not need to pretrain
on a different dataset, as in~\cite{Maqueda:2018}.
Biases were initialized to zero. 
Models were trained using a weight decay of $10^{-4}$ using the Adam optimizer~\cite{Kingma:2014} with 
an initial learning rate of $10^{-3}$. The networks were trained for 200 epochs, using minibatches of
128 samples and using a Mean Squared Error (MSE) loss.
Training time for one run on the prepared dataset using one NVIDIA K80 GPU took about 24 hours.

\subsection{Prediction of steering wheel angle} 
\label{subsec:steeringpredictionresults}

Fig.~\ref{fig:resnet:result} shows an example of the steering angle prediction. 
The top row shows example APS and DVS images.
The bottom plot compares the ground truth steering wheel angle
with the APS, DVS, and DVS+APS prediction results. 
While the three prediction curves appear similar, 
the prediction made by DVS+APS (in brown) is slightly more accurate. 
The dataset web site includes videos comparing predictions
from DVS, APS, and DVS+APS for all paper dataset recordings.

Table~\ref{tab:result}
summarizes  the RMS steering wheel angle prediction error (\textbf{RMSE}) and 
standard deviation achieved for each dataset using 
the combination of DVS and APS input channels. 
The standard deviation was computed over 5 repeats of each experiment, 
each time using different random seeds for weight initialization. 

The explained variance (\textbf{EVA}) also measures prediction accuracy. 
The EVA of steering angle $\Omega$, is defined by
${\rm EVA}=1-{\text{Var}(\vec{\Omega}_{\rm pred}-\vec{\Omega}_{\rm gt})/\text{Var}(\vec{\Omega}_{\rm gt})}$,
where $\vec{\Omega}_{\rm pred}$ and $\vec{\Omega}_{\rm gt}$ are 
the predicted and ground truth angles for all samples in the test set. 
EVA is dimensionless and ranges from approximately 0 to 1. The closer to 1, the better the prediction.

Both RMSE and EVA are useful for understanding the results. 
For example, the EVA for straight driving is usually quite low, 
indicating that the detailed timing of the
small steering corrections when driving straight are not very well predicted.
Since 
the RMSE is also small in this case, 
it means that prediction poorly reflects the details of steering but still 
well-predicts small angles. 
(During straight driving, humans only occasionally correct off-center lane positions,
and it is impossible to predict---especially from single frames---when these corrections occur.)

\begin{table}[t]
	\centering
    \footnotesize
    \caption{Steering wheel prediction error using RMSE and EVA. The colored bar shows the results of three datasets. 
    }\label{tab:result}
    \begin{tabular}{cccc}
    	\hline
        \multirow{2}{*}{\textbf{Dataset}} & \multicolumn{3}{c}{\textbf{RMSE ($^\circ$) (lower is better)}} \\
        \cline{2-4}
    	& \textbf{DVS+APS} & \textbf{DVS} & \textbf{APS} \\
		\hline
      	\texttt{Night} & $\mathbf{2.79\pm0.15}$ & $4.17\pm0.16$ & $3.49\pm0.06$ \\
        \texttt{Day} & $\mathbf{5.48\pm0.44}$ & $7.77\pm0.68$ & $7.30\pm0.38$\\
        \texttt{All} & $\mathbf{4.13\pm0.24}$ & $6.53\pm0.34$ & $5.60\pm0.25$\\
        \hline
        & \multicolumn{3}{c}{\textbf{EVA (higher is better)}} \\
        \cline{2-4}
    	& \textbf{DVS+APS} & \textbf{DVS} & \textbf{APS} \\
        \hline
      	\texttt{Night} & $\mathbf{0.931\pm0.008}$ & $0.851\pm0.011$ & $0.893\pm0.005$\\
        \texttt{Day} & $\mathbf{0.760\pm0.022}$ & $0.487\pm0.091$ & $0.537\pm0.049$ \\
        \texttt{All} & $\mathbf{0.881\pm0.009}$ & $0.668\pm0.032$ & $0.765\pm0.018$ \\
        \hline
    \end{tabular}\\
    \vspace{1mm}
    \includegraphics[width=0.9\linewidth]{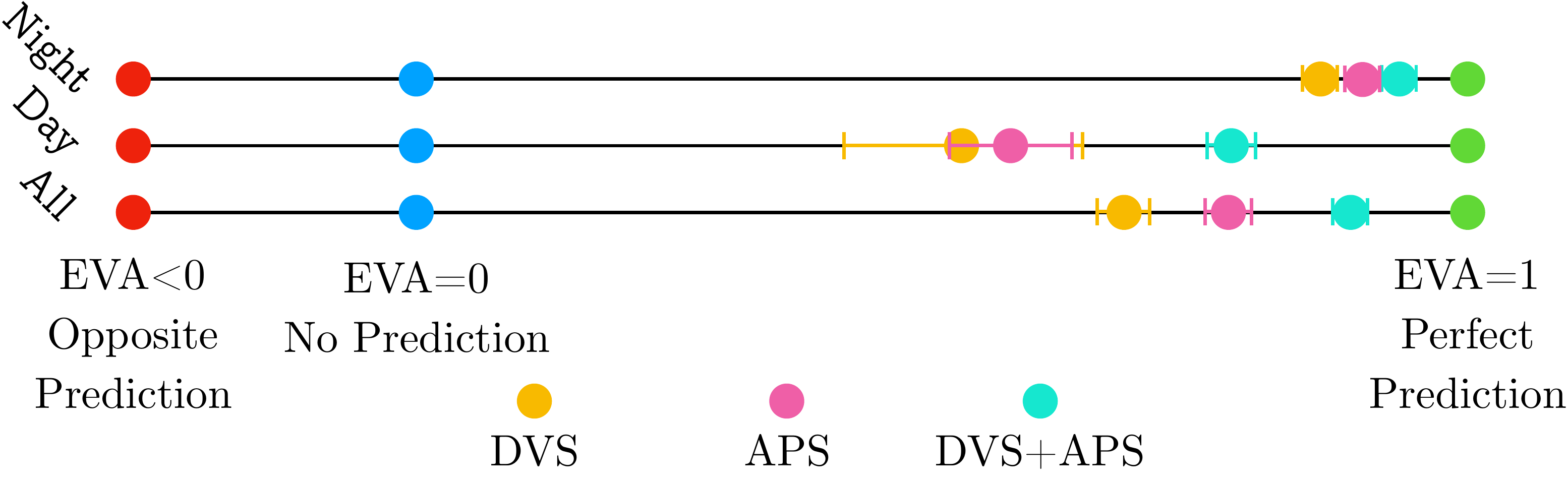}
\end{table}

The bold quantities in Table~\ref{tab:result}  highlight the best steering prediction
results for each dataset. The EVAs are compared graphically below the table.
It is clear that using DVS+APS  results in the best steering prediction.
A trivial baseline prediction error
 is obtained by fixing the prediction to
to $0^\circ$ (always driving straight).
This null prediction yields RMSE of $11^\circ$ and EVA of $0$, which also 
corresponds to the standard 
deviation of steering angle.

Even the worst RMSE of 7.8$^{\circ}$ and EVA of 0.49 obtained using only DVS in daylight conditions
are better than the null prediction.
Over all three datasets, the DVS+APS models achieve the best average steering 
angle prediction (EVA 0.88), which are slightly better 
than the best results obtained by~\cite{Maqueda:2018} (0.83). 
(Additionally, \cite{Maqueda:2018} used a larger ResNet-50,
and tested on interleaved time segments from each recording; 
i.e. the training used 40\,s sections of road immediately surrounding 20\,s testing sections).    
The EVA for DVS+APS is significantly better than for either DVS or APS alone.
It seems that the moving features
exposed by the DVS improve the steering predictions.
Overall, these results indicate that
the combined DVS and APS inputs help 
the network make more accurate predictions. 

Our overall EVA of $0.67\pm.03$ for \mbox{DVS-only} prediction are consistent with~\cite{Maqueda:2018}, who obtained overall EVA of 0.72.
But in sharp contrast,
we found that APS frames consistently produced better EVA of $0.77\pm.02$, while~\cite{Maqueda:2018} 
reported an overall \mbox{APS-only} EVA of only 0.40.
A higher EVA from APS frames might be expected, 
because the 
finer gray scale would generally allow seeing
road versus non-road more clearly. 

The nighttime steering predictions (EVA 0.93) are 
clearly better than daytime predictions (EVA 0.76).
We believe it is because the headlights do well to light the roadway 
to allow perception of the upcoming curves.
For daytime conditions, the fused APS+DVS provides a large improvement
of the EVA; it is more than 40\% higher with fused input than with either of the single inputs. 
We believe that it is from from 
the DVS better handling glare and overexposure.

\section{CONCLUSIONS}

DDD20 is 
the first open E2E driving dataset with 
over 50\,h of recordings from a DAVIS event camera mounted on a vehicle driven over 4000\,km.
The dataset increases the size of the original DDD17 dataset by a factor of 
about 4$\times$. In terms of both driving duration and distance, 
DDD20 is comparable to the 
72h NVIDIA dataset used in~\cite{Bojarski:2016} 
and the 10000\,km Baidu dataset~\cite{Yu:2017}. (The BDD100k dataset~\cite{yu_bdd100k:_2018} is far larger (1100h), but it is not E2E.)

We show the first results on end-to-end steering prediction
with the fused APS and DVS sensor input. 
Our results show that fused DVS and APS information best explains
the steering variance
under all driving conditions. 
Without temporal context, the DVS 
is blind to 
non-moving features that are common in driving, 
but provides valuable information to improve APS prediction.

Future work could exploit the fine timing of the DVS events, for example, to preprocess the input for optical flow, to compute inference on DVS interframes, and to incorporate temporal context into the predictions. This temporal context could better enable prediction of throttle and braking decisions that are difficult  using single frames.

\addtolength{\textheight}{-12cm}   





\section*{ACKNOWLEDGMENT}

We thank  D.\,Rettig, A.\,Stockklauser, G.\,Detorakis, G.\,Burman, and D-D.\,Delbruck for co-piloting; J.\,Anumula for help with data analysis; \href{https://inivation.com/}{iniVation} and the \href{http://sensors.ini.uzh.ch}{INI Sensors Group} for device support. We are especially grateful to the 2017 Telluride Neuromorphic Engineering Workshop which provided the opportunity for the DDD20 data collection.

\bibliographystyle{IEEEtran}
\bibliography{paperef}

\end{document}